\documentclass[10pt,twocolumn,letterpaper]{article}

\usepackage{wacv}
\usepackage{times}
\usepackage{epsfig}
\usepackage{graphicx}
\usepackage{amsmath}
\usepackage{amssymb}

\usepackage{algorithm}
\usepackage{algpseudocode}
\usepackage{color, colortbl}
\usepackage{graphicx}
\usepackage{subcaption}
\usepackage{booktabs}  
\usepackage[flushleft]{threeparttable}

\definecolor{Gray}{gray}{0.9}

\wacvfinalcopy 

\def\wacvPaperID{27} 

\ifwacvfinal\fi
\begin{document}

\title{Learning to Prune Filters in Convolutional Neural Networks}


\author{Qiangui Huang$^1$
 \qquad Kevin Zhou$^2$ \qquad Suya You$^3$ \qquad Ulrich Neumann$^1$\\
$^1$University of Southern California \hspace{5mm}  $^2$ Siemens Healthineers \hspace{5mm}  $^3$US Army Research Laboratory\\
\hspace{2mm}Los Angeles, California \hspace{12mm} Princeton, New Jersey \hspace{10mm} Playa Vista, California\\
\hspace{-5mm}{\tt\small \{qianguih,uneumann\}@usc.edu}\hspace{10mm}  {\tt\small s.kevin.zhou@gmail.com}\hspace{10mm}  {\tt\small suya.you.civ@mail.mil}\qquad
}


\maketitle
\ifwacvfinal\thispagestyle{empty}\fi

\begin{abstract}
Many state-of-the-art computer vision algorithms use large scale convolutional neural networks (CNNs) as basic building blocks. These CNNs are known for their huge number of parameters, high redundancy in weights, and tremendous computing resource consumptions. This paper presents a learning algorithm to simplify and speed up these CNNs. Specifically, we introduce a ``try-and-learn" algorithm to train pruning agents that remove unnecessary CNN filters in a data-driven way. With the help of a novel reward function, our agents removes a significant number of filters in CNNs while maintaining performance at a desired level. Moreover, this method provides an easy control of the tradeoff between network performance and its scale. Performance of our algorithm is validated with comprehensive pruning experiments on several popular CNNs for visual recognition and semantic segmentation tasks. 
\end{abstract}


\section{Introduction}
Modern computer vision tasks rely heavily on carefully designed convolutional neural networks architectures \cite{alexnet, vgg, googlenet, resnet}. These CNNs usually consist of multiple convolutional layers with a large amount of parameters. They are computationally expensive and are often over-parametrized \cite{ppdl}. Recently, network pruning has become an important topic that aims at simplifying and accelerating large CNN networks.

In  \cite{han_both,han_deep} , Han  \textit{et al.} proposed compressing CNNs by removing weights with small magnitudes and achieved promising compression results. Pruning individual weights increases the sparsity in CNNs. In order to get real compression and speedup, it requires specially designed software (like sparse BLAS library) or hardware \cite{eie} to handle the sparsity. Pruning filters is another means to simplify CNNs. An advantage of pruning filters is that it provides both compression and speedup benefits without requiring specially designed software or hardware. Moreover, pruning filters can be used in addition to other sparsity or low-rank-approximation based methods to further reduce computations. In \cite{iclr17}, Hao \textit{et al.} pruned filters according to a hand-crafted criteria, the L1 norm. They showed that pruning filters with small L1 norms gives better pruning performance than random pruning or pruning the largest filters. However, it remains unknown that if pruning filters in a data-driven way offers better pruning performances. 

Moreover, in many practical scenarios, it is desirable to have an easy control of the tradeoff between network performance and scale during pruning. However, to the best of our knowledge, this is not available in existing works. For example, in some situations, we are willing to sacrifice certain level of performances. Unfortunately, it usually takes tremendous human efforts to test different tradeoffs between the network performance and scale during pruning in order to achieve the best pruning performance under the desired performance drop budget.


Our work focuses on addressing aforementioned problems. Firstly, our method learns to prune redundant filters in a data-driven way. We show that pruning in a data-driven way gives better performances than the hand-crafted pruning criteria \cite{iclr17}. Secondly, our method provides an easy control of the tradeoff between network performance and scale during pruning. After specifying the desired network performance, our method automatically outputs a compact model with filters aggressively pruned without involving humans in the loop.

In order to achieve these goals, we formulate the filter pruning problem as a ``try-and-learn" learning task. Specifically, we train a pruning agent, modeled by a neural network, to take the filter weights as input and output binary decisions to remove or keep filters. The agent is trained with a novel reward function which encourages high pruning ratios and guarantees the pruned network performance remains above a specified level. In another word, this reward function provides an easy control of the tradeoff between network performance and scale. Since the reward function is non-differentiable w.r.t the parameters of pruning agents, we use the policy gradient method \cite{pg, reinforce} to update the parameters in training.

Intuitively, our algorithm functions as a ``try-and-learn" process. The pruning agent starts by guessing which filters to prune. Every time it takes a pruning action, the action is evaluated by our reward function. The reward is then fed back to the agent which supervises the agent to output actions with higher rewards. Often the search space is extremely large. For example, for a layer with 64 filters, there are $2^{64}$ different options of which filters to remove. However, our algorithm is highly efficient and experiments show that it converges after only a relatively small number of trials. 

Our method is totally automatic and data-driven. Once started, it automatically discovers redundant filters and removes them while keeping the performance of the model within a specified tolerance. Our method also makes it possible to control the tradeoff between network performance and its scale during pruning without involving humans in the loop. We demonstrate the performance of our algorithm through extensive experiments with various CNN networks in section \ref{sec:exp}.

\section{Related Works}

Early works on model compression are based on the saliency guidance \cite{optimal} or the Hessian of the loss function \cite{second}. Recently, there are many pruning methods proposed for modern large scale CNNs. We review related works in the following categories.

\textbf{Low-rank approximation}. LRA methods \cite{exploit_linear,lr_expansion} are based on one key observation that most of CNN filters or features are of low rank and can be decomposed into to lightweight layers by matrix factorization. \cite{exploit_linear} made one of the early attempts at applying LRA methods such as Single Value Decomposition (SVD) for network simplification. \cite{lr_expansion} decomposed $k \times k$ filters into $k \times 1$ and $1 \times k$ filters, which effectively saved the computations. \cite{lr_expansion} investigated two different optimization schemes, one for filter-based approximation and one for feature-based approximation. Similarly, \cite{lccl} used low-cost collaborative kernels for acceleration. \cite{zhang_accelerating} used Generalized SVD for the non-linearity in networks and achieved promising results in very deep CNNs. Recently, \cite{yihuihe} combined the low-rank approximation with channel pruning and \cite{Wen_2017_ICCV} proposed to use Force Regularization to train neural networks towards low-rank spaces.

\textbf{Increasing Sparsity}. Methods in this category mainly focus on increasing the sparsity in CNNs. \cite{han_both,han_deep} introduced a three-step training pipeline to convert a dense network into a sparse one. They achieved this by removing connections with small weights. Their method showed promising compression rates on various CNNs. However, it required an additional mask to mask out pruned parameters and handle the sparsity, which actually does not save computations. \cite{dns} proposed a dynamic network surgery algorithm to compress CNNs by making them more sparse. \cite{alvarez2016learning,zhou2016less,BMVC2015_31} also proposed to remove the neurons in networks to increase the sparsity. Usually, these methods require specially designed software or hardware \cite{eie} to handle the sparsity in CNNs in order to gain real speedup.

\textbf{Quantization and binarization}. On top of the method in \cite{han_both}, \cite{han_deep} used additional quantization and Huffman encoding to further compress the storage requirement. These methods can also be applied on top of our filter pruning algorithm. Another group of methods tried to use bit-wise operations in CNNs to reduce the computations. \cite{bnn} introduced an algorithm to train networks with binary weights and activations. In \cite{xnor}, the authors suggested to also binarize the inputs to save more computations and memories. However, experiments showed that the performance of these binarized networks are worse than their full prevision counterparts \cite{xnor}.

\textbf{Pruning filters}. Pruning filters has some nice properties as mentioned above. In \cite{iclr17}, the authors proposed a magnitude-based pruning method. Our method differs from their work in 1). our algorithm is able to prune filters in a data-driven way which gives better pruning performances as shown in section \ref{sec:exp}; 2). our algorithm supports the control of the tradeoff between network performance and scale without involving humans in the loop.

\textbf{Reinforcement learning}. There are some works applying reinforcement learning methods to neural network architecture design. In \cite{ns}, the authors proposed to generate neural network architecture descriptions using a recurrent neural network (RNN). They trained the RNN via policy gradient method. \cite{nd} explored with more possible architectures using Q-learning \cite{Q1, Q2}. These methods are all used to generate neural network architectures from scratch and require tremendous computing resources for training. In contrast, our work focuses on pruning a given baseline network and consumes much less computing resources.

\begin{algorithm}[!ht]
  \caption{Prune filters of one layer in CNN}\label{algo1}
  \begin{algorithmic}
    
  \State \textbf{Input}: a baseline model $f$, the index of layer to prune $l$, $\mathbf{X}_{train}$, $\mathbf{X}_{val}$, learning rate $\alpha$
  \State \textbf{Output}: the parameters of pruning agent $\theta^l$, a pruned network $\hat{f}$
  
       \While {not converged} 
           \State Zero all gradients $\nabla_{\theta^l}\mathcal{L} \leftarrow 0 $
           \State Initialize an action-reward buffer $AR$
           \For{i=1,2,..., M}                                                                     \Comment{Sample the output distribution $M$ times}   
               \State $\mathbf{A_i^l}  \sim \pi^l(W^l, \theta^l)$                                       
               \State Surgery $f$ according to $\mathbf{A_i^l} $ and get a new model $\hat{f}_{\mathbf{A_i^l} }$                 
               \State Fine tune $\hat{f}_{\mathbf{A_i^l} }$  using $\mathbf{X}_{train}$
               \State Evaluate $\hat{f}_{\mathbf{A_i^l} }$ and calculate the reward $R(\mathbf{A_i^l} )$ by equation (1)
               \State Store the action $\mathbf{A_i^l} $ and reward $R(\mathbf{A_i^l} )$ pair in the buffer $AR$
          \EndFor
          \State Normalize the rewards stored in the buffer $AR$
           \For{i=1,2,..., M} 
              \State Retrieve the $i^{th}$ action and reward pair
              \State Calculate the gradient: $\nabla_{\theta^l}\mathcal{L} \mathrel{+}= R(\mathbf{A_i^l} ) * \nabla_{\theta^l} \log(\mathbf{A_i^l}   | \pi^l(W^l,\theta^l))  $
          \EndFor      
        \State Update the parameters: $\theta^l \leftarrow \theta^l + \alpha * \nabla_{\theta^l}\mathcal{L}$
     \EndWhile
     \State prune $f$ by the agent $\pi^l$ and output pruned network $\hat{f}$
  \end{algorithmic}
\end{algorithm}

\section{Method}

We introduce the proposed method for pruning a CNN in this section. Assume there is a baseline CNN $f$ with $L$ convolutional layers. Firstly, we describe how to prune an individual layer. Then, we present details about how to prune all layers in a CNN.

\subsection{Learning to prune filters in one individual layer}
Let $N_l$ denote the number of filters in the $l^{th}$ layer in the baseline CNN $f$. Suppose the filter matrix is represented by $W^l = \{w^l_{1}, w^l_{2}, ..., w^l_{i},..., w^l_{N_l}\}$, where $w^l_{i} \in \mathbb{R}^{m^l \times h \times w}$ with $m^l$, $h$, and $w$ being the number of input feature maps, the height and width of the $l^{th}$ layer filter. Our goal is to simplify the network by pruning redundant filters in $W^l$. 

We achieve this by training a pruning agent $\pi^l$ which takes $W^l$ as the input and makes a set of binary actions $\mathbf{A^l} = \{ a^l_1, a^l_2, ..., a^l_i, ..., a^l_{N^l}  \}$. Here $a^l_i \in \{0,1\}$. $a^l_i = 0$ means the agent treats the $i^{th}$ filter as unnecessary and decides to remove it and $a^l_i = 1$ means the opposite. The agent is parametrized by $\theta^l$. Formally, the agent models the following conditional probability, $\pi^l( \mathbf{A^l} | W^l, \theta^l  )$. We use a neural network to model the pruning agent in our experiments.

For the given task, there is a validation set $\mathbf{X}_{val} = \{  \mathbf{x}_{val}, \mathbf{y}_{val}  \}$. $\mathbf{x}_{val}$ and $\mathbf{y}_{val}$ denote the set of validation images and the corresponding ground truth, respectively. The learning goal is to maximize the objective $\mathcal{L} = R( \mathbf{A^l}, \mathbf{X}_{val} )$. Here $R$ is a reward function defined as the multiplication of two terms, the accuracy term and efficiency term.

 \begin{gather}
  R(\mathbf{A^l}, \mathbf{X}_{val}) = \psi(\mathbf{A^l}, \mathbf{X}_{val}, b, p^*) * \varphi(\mathbf{A^l})
\end{gather}

The \textbf{accuracy term} $\psi(\mathbf{A^l}, \mathbf{X}_{val}, b, p^*)$ is calculated by equation (2). It guarantees the performance drop evaluated on $\mathbf{X}_{val}$ under metric $\mathcal{M}$ is bounded by $b$. The performance drop bound $b$ is a hyper-parameter used to control the tradeoff between network performance and scale. $\hat{p}$ and $p^*$ are the performance of new model $\hat{f}_{\mathbf{A^l}}$ and baseline model $f$. The new model $\hat{f}_{\mathbf{A^l}}$ is generated by surgerying $f$ according to the action set $\mathbf{A^l}$. Before evaluation, $\hat{f}_{\mathbf{A^l}}$ is first fine-tuned by a training set $\mathbf{X}_{train} = \{  \mathbf{x}_{train}, \mathbf{y}_{train}  \}$ for some epochs to adjust to the pruning actions. In evaluation, the metric $\mathcal{M}$ is set as the classification accuracy in recognition tasks and the global accuracy in segmentation tasks in this paper. The larger $\hat{p}$ is, the more $\psi$ contributes to the final reward. However, when the performance drop $(p^* - \hat{p})$ is larger than the bound $b$, $\psi$ contributes negatively to the final reward. This forces the pruning agent to keep the performance of pruned network above a specified level.

 \begin{gather}
    \psi(\mathbf{A^l}, \mathbf{X}_{val}, b, p^*) = \frac{ b - (p^* - \hat{p})}{b} \\
    \hat{p} = \mathcal{M}(\hat{f}_{\mathbf{A^l}},\mathbf{X}_{val}), \quad  p^* = \mathcal{M}(f, \mathbf{X}_{val})
\end{gather}

The \textbf{efficiency term} $\varphi(\mathbf{A^l})$ is calculated by equation (4). It encourages the agent $\pi^l$ to prune more filters away. $\mathcal{C}(\mathbf{A^l}) $ denotes the number of 1-actions in $\mathbf{A^l}$ which is also the number of kept filters. A small $\mathcal{C}(\mathbf{A^l}) $ means only a few filters are kept and most of the filters are removed. The smaller $\mathcal{C}(\mathbf{A^l}) $ is, the more efficient the pruned model is, and more $\varphi$ contributes to the final reward. The $\log$ operator guarantees two terms in equation (1) are of the same order of magnitude.

 \begin{gather}
   \varphi(\mathbf{A^l}) = \log( \frac{N^l}{ \mathcal{C}(\mathbf{A^l}) } )
\end{gather}

Since $\mathcal{L}$ is non-differentiable w.r.t $\theta^l$, we use the policy gradient estimation method \cite{pg}, specifically the REINFORCE \cite{reinforce}, to estimate the gradients $\nabla_{\theta^l}\mathcal{L}$ as equation (5). Furthermore, we can sample $M$ times from the output distribution to approximate the gradients. This gives the final gradient estimation formula in equation (6). In order to get an unbiased estimation, the rewards of $M$ samples are normalized to zero mean and unit standard deviation. Algorithm \ref{algo1} summarizes the whole training process .

 \begin{gather}
    \nabla_{\theta^l}\mathcal{L} = \mathbb{E}[R(\mathbf{A^l}) * \nabla_{\theta^l} \log(\pi^l(\mathbf{A^l}|W^l, \theta^l))] \\
    \nabla_{\theta^l}\mathcal{L} = \sum_{i=1}^{M}[ R(\mathbf{A_i^l} ) * \nabla_{\theta^l} \log( \mathbf{A_i^l}  | \pi^l(W^l,\theta^l)) ]
\end{gather}


\begin{algorithm}[!ht]
  \caption{Prune filters in the entire CNN}\label{algo2}
  \begin{algorithmic}
    
  \State \textbf{Input}: a baseline model $f$ with $L$ convolutional layers to prune, $\mathbf{X}_{train}$, $\mathbf{X}_{val}$, learning rate $\alpha$
  \State \textbf{Output}: the parameters of pruning agents $\boldsymbol{\theta} = \{ \theta^1, \theta^2, ..., \theta^l, ..., \theta^L   \}$, a pruned model $\hat{f}$
  \State Initialize the pruned model: $\hat{f} \leftarrow f$ 
  \For{l=1,2,..., L} 
      \State initialize the pruning agent $\pi^l$
      \State do the training of pruning agent $\pi^l$ using Algorithm \ref{algo1} with input $\hat{f}$, $l$, $\mathbf{X}_{train}$, $\mathbf{X}_{val}$
      \State prune $f$ by the agent $\pi^l$ and updates $\hat{f}$
      \State fine tune $\hat{f}$ using $\mathbf{X}_{train}$ to compensate performance decrease
  \EndFor   
  \end{algorithmic}
\end{algorithm}

\begin{table*}
  \caption{Statistics of baseline models.}
  \label{tl:baseline}
  \centering
  \begin{tabular}{llllll}
    \toprule

    Method     		   & parameters   & Number of FLOPs   & GPU Time (s) & CPU Time (s)  & Accuracy (\%) \\
    \midrule
    
    VGG-16 	   &  15M	& 3.1E+08 & 0.138  &8.5 & 92.77  \\
    ResNet-18 	   &  11M & 1.3E+10 & 0.226	&13.6& 93.52	\\
    FCN-32s	   & 136M  & 7.4E+10 & 0.115	&  2.5  & 90.48 		\\
    SegNet 	   &  	29M & 4.0E+10 & 0.156	& 4.3 & 	86.50	\\    

    \bottomrule
  \end{tabular}

\end{table*}

\begin{figure*}[h]
  \centering
  \captionsetup{justification=centering}
  \begin{subfigure}{.5\textwidth}
  \centering
  \includegraphics[width=0.7\linewidth]{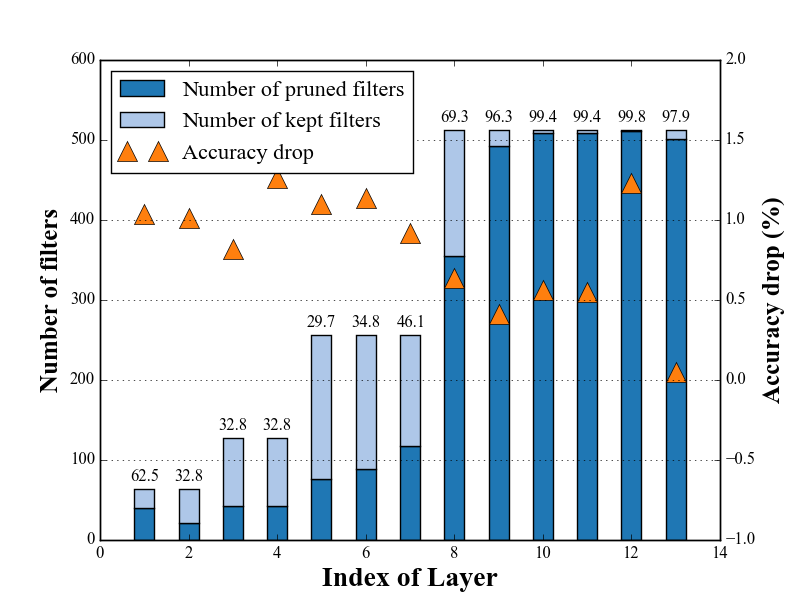}
   \captionof{figure}{Results of pruning single layer in VGG-16 on CIFAR-10}
  \label{fig:vgg_cifar_in}
\end{subfigure}%
\hfill
\begin{subfigure}{.5\textwidth}
  \centering
  \includegraphics[width=0.7\linewidth]{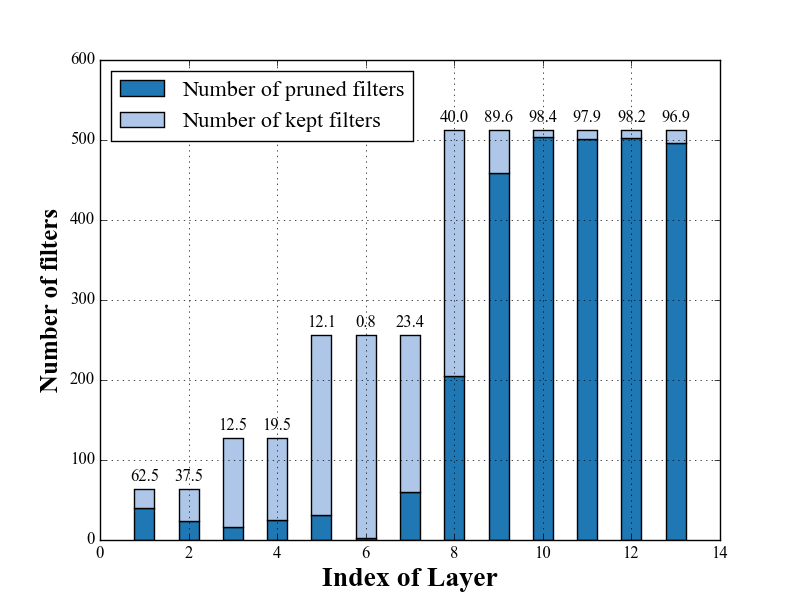}
   \captionof{figure}{Results of pruning all layers in VGG-16 on CIFAR-10}
  \label{fig:vgg_cifar_all}
\end{subfigure}%
\caption{ Pruning results of VGG-16 on CIFAR 10. Numbers on top of bars are the pruning ratios.}
\label{fig:vgg_cifar}
\end{figure*}

\subsection{Learning to prune filters in all layers in a network}

In a baseline with $L$ convolutional layers, our algorithm prunes all of them by training a set of pruning agents $\boldsymbol{\pi} = \{ \pi^1, \pi^2, ..., \pi^l, ..., \pi^L   \}$ where $\pi^l$ prunes the filters in the $l^{th}$ layer. It starts from lower layers and proceeds to higher layers one by one. Algorithm \ref{algo2} summarizes the overall training process. After finishing the pruning of one layer, it fine-tunes the entire network for some epochs again using $\mathbf{X}_{train}$ to compensate the performance decrease. We find this layer-by-layer and low-to-high pruning strategy works better than other alternatives such as pruning all layers by their order of pruning-sensitivity.

\begin{figure}
\begin{center}
   \includegraphics[width=0.7\linewidth]{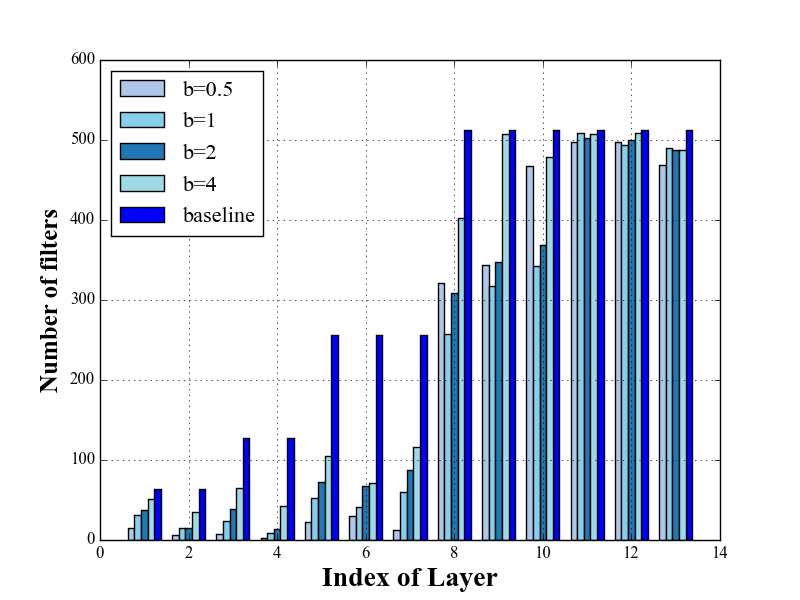}
\end{center}
   \caption{Result comparison of pruning the VGG-16 network on CIFAR 10 using different drop bounds.}
\label{fig:vgg_cifar_in}
\end{figure}

\begin{figure}
\begin{center}
   \includegraphics[width=0.7\linewidth]{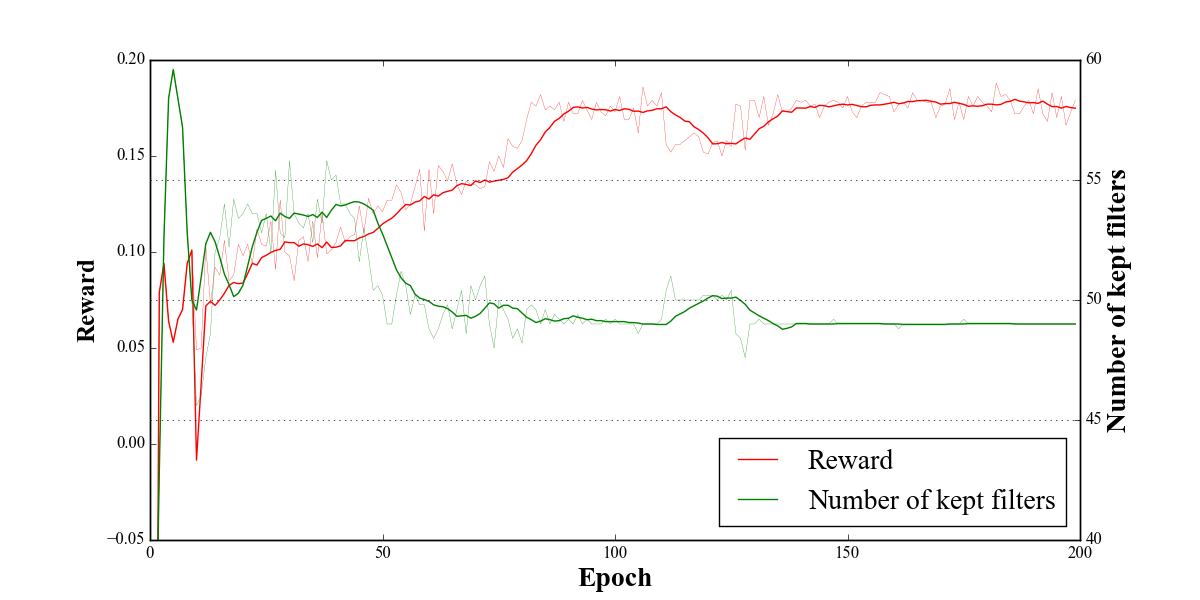}
\end{center}
   \caption{Training curves of the pruning agent for the $2^{nd}$ layer in VGG-16 on CIFAR 10. Originally there are 64 filters. Red line is the reward. Green line denotes the number of kept filters. Light lines are the raw data and bold lines are the smoothed data for better visualization.}
\label{fig:tc}
\end{figure}

\begin{figure}[h]
  \centering
  \captionsetup{justification=centering}
  \centering
  \includegraphics[width=0.7\linewidth]{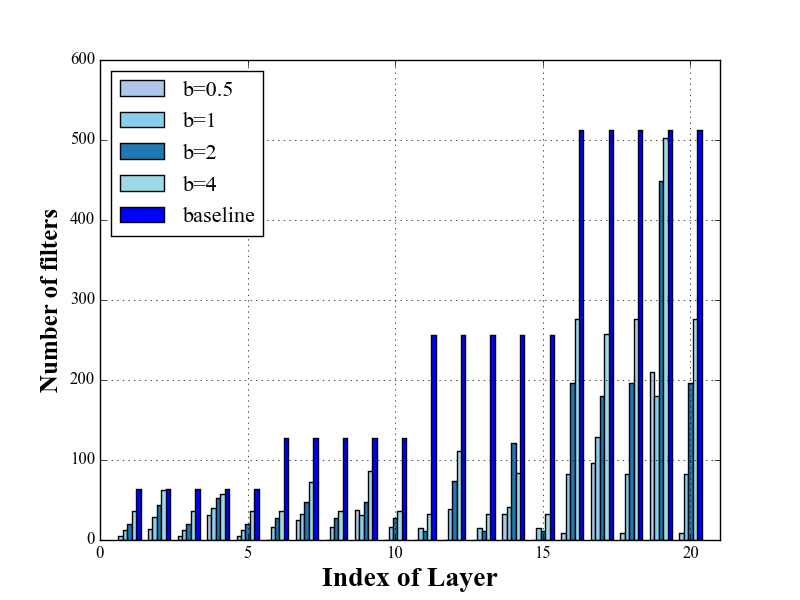}
  \label{fig:vgg_cifar_in}
   \caption{Result comparison of pruning the ResNet-18 network on CIFAR 10 using different drop bounds.}
\label{fig:resnet}
\end{figure}

\begin{table*}[t]
  \caption{Various pruning results of VGG-16 network on CIFAR 10.}
  \label{tl:vgg_cifar}
  \centering
  \begin{tabular}{llllll}
    \toprule

    Method     & Prune Ratio (\%)  & Saved FLOPs (\%)  & GPU Speedup (\%) &  CPU Speedup (\%) & Accuracy drop \\
    \midrule
    
    ICLR 17 \cite{iclr17}  & 64.0  &  34.0	& 31.4 &32.4& -0.1   \\
    \hline
    Ours $b$ = 0.5 	   & 83.3	& 45.0 & 32.4	& 48.8 &  \textbf{0.6} (1.0)	\\ 
    Ours $b$ = 1 	   &  82.7	& 55.2  & 37.8	&56.6& \textbf{1.1} (2.0)	\\
    Ours $b$ = 2 	   & 86.5	& 64.5 & 44.6	&63.4& \textbf{1.9} (2.4)		\\
    Ours $b$ = 4 	   & 92.8	& 80.6 & 63.4	&79.2& \textbf{3.4} (6.4)		\\    
    \bottomrule
  \end{tabular}
  
      \begin{tablenotes}
      \small
      \item $\star$ Note that numbers in parenthesis denote the accuracy drop of the method in \cite{iclr17} applied to the same baseline model with same pruning ratios.
    \end{tablenotes}
    
\end{table*}

\section{Experiments}
\label{sec:exp}

We experiment with the pruning algorithm on two tasks, visual recognition and semantic segmentation. For the visual recognition task, we test the VGG-16 \cite{vgg} and ResNet-18 \cite{resnet} network on the CIFAR 10 dataset \cite{cifar}. For the segmentation task, we test the FCN-32s network \cite{fcn} on Pascal VOC dataset \cite{pascal} and the SegNet network \cite{segnet} on the CamVid dataset \cite{camvid}. Pruning agents in all experiments are modeled by neural networks built by following the same protocol. For VGG-16 and ResNet-18 network, we use the official PyTorch implementation \footnote{The first max pooling layer in the official ResNet-18 network is removed to deal with the small resolution.}. For the FCN-32s and SegNet network, we use their official caffe \cite{caffe} implementation. The performances of baseline models are reported in Table\ref{tl:baseline}. For all datasets, we select a hold-out set from training set as our validation set.

\textbf{Pruning agent design protocol}. All pruning agents are modeled by neural networks which takes the filter matrix  $W^l$ (of size $N_l \times m^l \times h \times w$) as input and outputs $N_l$ binary decisions. In order to apply same operations to all filter weights, $W^l $ is first rearranged into a 2D matrix of size $N_l \times M^l$ where $M^l=m^l \times h \times w$ and then fed into the pruning agent. If $M^l$ is larger than $2^4$, the pruning agent will be composed of 4 alternating convolutional layers with $7\times7$ kernels and pooling layers followed by two fully connected layers. Otherwise, the pruning agent will only consist of two fully connected layers. In training, all pruning agents are updated with the Adam optimizer \cite{adam}. We use a fixed learning rate 0.01 for all recognition experiments and 0.001 for all segmentation experiements. We roll out the output distributions for 5 times ($M=5$) for all experiments.

\textbf{Distributed training}. The output distribution sampling and evaluation process in Algorithm \ref{algo1} is the most time-consuming part in our algorithm. However, it is highly parallelizable and we use distributed computation in experiments for speed up. All training are conducted on Nvidia K40 GPUs. GPU speeds are measured on one single K40 while CPU speeds are measured on one core of a Xeon(R) CPU E5-2640 v4 CPU. All measurements are averaged over 50 runs. The measurements use a batch of 512 $32\times32$ images for recognition networks, a batch of one $256\times256$ image for the FCN-32s and a batch of one $360\times480$ image for the SegNet. The training process takes $\sim$5 days for pruning on CIFAR 10 networks and SegNet and $\sim$10 days for FCN-32s network using 6 GPUs.

\begin{table*}
  \caption{Various pruning results of ResNet-18 network on CIFAR 10.}
  \label{tl:resnet_cifar}
  \centering
  \begin{tabular}{llllll}
    \toprule

    Method     & Prune Ratio (\%)  & Saved FLOPs (\%)   & GPU Speedup (\%) & CPU Speedup (\%) & Accuracy drop \\
    \midrule
    
    Ours $b$ = 0.5 	   &  27.1	& 24.3 & 6.2	&26.9&  \textbf{0.3} (1.3)	\\

    Ours $b$ = 1 	   &  37.0	& 35.3 &13.1 	&42.0& \textbf{1.0} (6.9)	\\

    Ours $b$ = 2 	   &  67.9  & 64.7 & 31.1	&60.0& \textbf{1.7} (6.5)		\\

    Ours $b$ = 4 	   &  78.4	& 76.0 & 54.7	&74.2& \textbf{2.9} (21.0)		\\    

    \bottomrule
  \end{tabular}
  
        \begin{tablenotes}
      \small
      \item $\star$ Note that numbers in parenthesis denote the accuracy drop of the method in \cite{iclr17} applied to the same baseline model with same pruning ratios.
    \end{tablenotes}
    
\end{table*}

\subsection{Pruning VGG-16 on CIFAR 10}
\label{sec:vgg_cifar}

We first prune single layer in VGG-16 on CIFAR 10 using Algorithm \ref{algo1}. The accuracy drop bound $b$ is set as 2. The results are reported in Fig.\ref{fig:vgg_cifar} (a). As a byproduct, our algorithm can analyze the redundancy in each layer of a given CNN. Pruning results show that in VGG-16 on CIFAR-10, higher layers contain more unnecessary filters than lower layers as removing more than 95\% of the filters in some higher layers (layer 9, 10, 11, 13) has relatively less impact on performance. One set of typical training curves is presented in Fig. \ref{fig:tc}. As the number of training epochs increase, the reward keeps increasing and more and more filters are removed. Originally, there are 64 filters in this layer, which means there are $2^{64}$ different decision options. However, our agent converges after less than 200 epochs in this case. This shows the efficiency of our algorithm.

We now prune all layers in the VGG-16 network on CIFAR 10 using the same drop bound $b=2$. The pruning results are presented in Fig.\ref{fig:vgg_cifar} (b). Generally, the pruning ratio of each layer is slightly smaller than the single-layer pruning. But our algorithm still aggressively prunes lots of redundant filters. Several quantitative evaluations are reported in Table \ref{tl:vgg_cifar}. We also experiment with various drop bound $b$ to show how to control the tradeoff between network performance and scale using our algorithm. All the results are reported in Table \ref{tl:vgg_cifar}. Note that the final accuracy drops are usually not exactly same as the drop bound. That is because there is a generalization gap between validation and final test set.

\begin{figure}
  \centering
  \captionsetup{justification=centering}
  \includegraphics[width=0.9\linewidth]{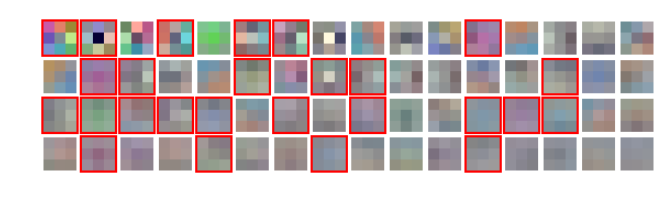}
\caption{Visualization of filters. Filters are ranked by its $L1$ norm in a descending order from top left to bottom right. Filters with red bounding boxes are the ones kept by our algorithm (with $b=2$).}
\label{fig:filter}
\end{figure}

For further comparisons, we also apply the magnitude-based filter pruning method in \cite{iclr17} to the same baseline network and prune the network using the same pruning ratios. The accuracy drops of these pruned networks are reported in the parenthesis in Table \ref{tl:vgg_cifar}. Two key observations include: 1). A larger drop bound generally gives a higher prune ratio, more saved FLOPs, a higher speedup ratio, and a larger accuracy drop on test set; 2). With the same pruning ratios, our data-driven algorithm discovers better pruning solutions (which results in less accuracy drops) for VGG-16 on CIFAR 10 than the method in \cite{iclr17}. For further comparison with \cite{iclr17}, we also visualize the filters in the $1^{st}$ layer in VGG-16 on CIFAR 10 in Fig.\ref{fig:filter}. They show that our algorithm does not prune filters based on their magnitude.

\subsection{Pruning ResNet-18 on CIFAR 10}

With promising results on VGG-16 on CIFAR 10, we also experiment with a larger and more complex network, ResNet-18 \cite{resnet}, on the same CIFAR 10 dataset. Different drop bounds are tested as well. Results are reported and compared with \cite{iclr17} \footnote{Note that in \cite{iclr17}, the authors only pruned the first layer in a residual block. Since our algorithm prunes other layers in a residual block, we also prune other layers using their method in the comparison experiments to achieve the same pruning ratio.} in the same way as section \ref{sec:vgg_cifar} in Table. \ref{tl:resnet_cifar}. Compared with VGG-16, ResNet-18 is a more lightweight and efficient network architecture. Thus, the overall pruning ratios are smaller. However, our algorithm is still capable to prune a lot of redundant filters.

In Fig.\ref{fig:resnet}, we show the detailed pruning ratios of each layer with different drop bounds. In total, there are 20 convolutional layers in the ResNet-18 network including the shortcut convolutional layers in residual blocks. Generally, larger drop bounds offer larger pruning ratios. Moreover, results show that in a residual block, the first convolution layer is easier to prune than the second one. For example, there are more filters are removed in the $7^{th}/9^{th}/12^{th}/14^{th}$ layer than the $8^{th}/10^{th}13^{th}/15^{th}$ layer, regardless of the drop bound.

\subsection{Pruning FCN-32s on Pascal VOC}

\begin{table*}[t]
  \caption{Pruning results of segmentation networks.}
  \label{tl:fcn}
  \centering
  \begin{tabular}{lllllll}
    \toprule

    Method     & Prune Ratio (\%)  & Saved FLOPs (\%)   & GPU Speedup (\%) & CPU Speedup (\%) & Accuracy drop \\
    \midrule
    
    FCN32s 	   &  63.7	& 55.4 & 37.0	&49.1    &  \textbf{1.3} (3.4)	\\
    
    SegNet 	   &  56.9	 & 63.9 & 42.4	&53.0 &  \textbf{-2.1} (3.0)	\\
    
    \bottomrule
  \end{tabular}
  
    \begin{tablenotes}
      \small
      \item $\star$ Note that numbers in parenthesis denote the accuracy drop of the method in \cite{iclr17} applied to the same baseline model with same pruning ratios.
    \end{tablenotes}
    
  
\end{table*}

CNNs designed for semantic segmentation tasks are much more challenging to prune as the pixel-level labeling process requires more weights and more representation capacities. In order to show the performance of our pruning algorithm on segmentation tasks, we apply our algorithm to a pre-trained FCN-32s network \cite{fcn} on the Pascal VOC dataset \cite{pascal}. We use the global pixel accuracy as the evaluation metric $\mathcal{M}$ in our reward function. Note that depending on the scenario, other metrics, such as per-class accuracy and mean IU, can also be used as metric $\mathcal{M}$. The drop bound $b$ is set as 2. 

Following Algorithm \ref{algo2}, our algorithm removes near 63.7\% redundant filters in FCN-32s and the inference process is accelerated by 37.0\% on GPU and 49.1\% on CPU as reported in Table. \ref{tl:fcn}. In the FCN-32s network, the last two convolutional layers are converted from fully connected layers (one is of size $512 \times 4096 \times 7 \times 7$ and the other one is of size $4096 \times 4096 \times 1 \times 1$). These two layers contribute 87.8\% of the parameters in the entire network and are of high redundancy. In our algorithm, 51.6\% and 68.7\% of the filters in these two layers are removed, respectively. Detailed pruning ratios of each layer in the FCN-32s network is presented in Fig. \ref{fig:ratio_fcn}. 

\begin{figure*}
\begin{center}
   \includegraphics[width=0.8\linewidth]{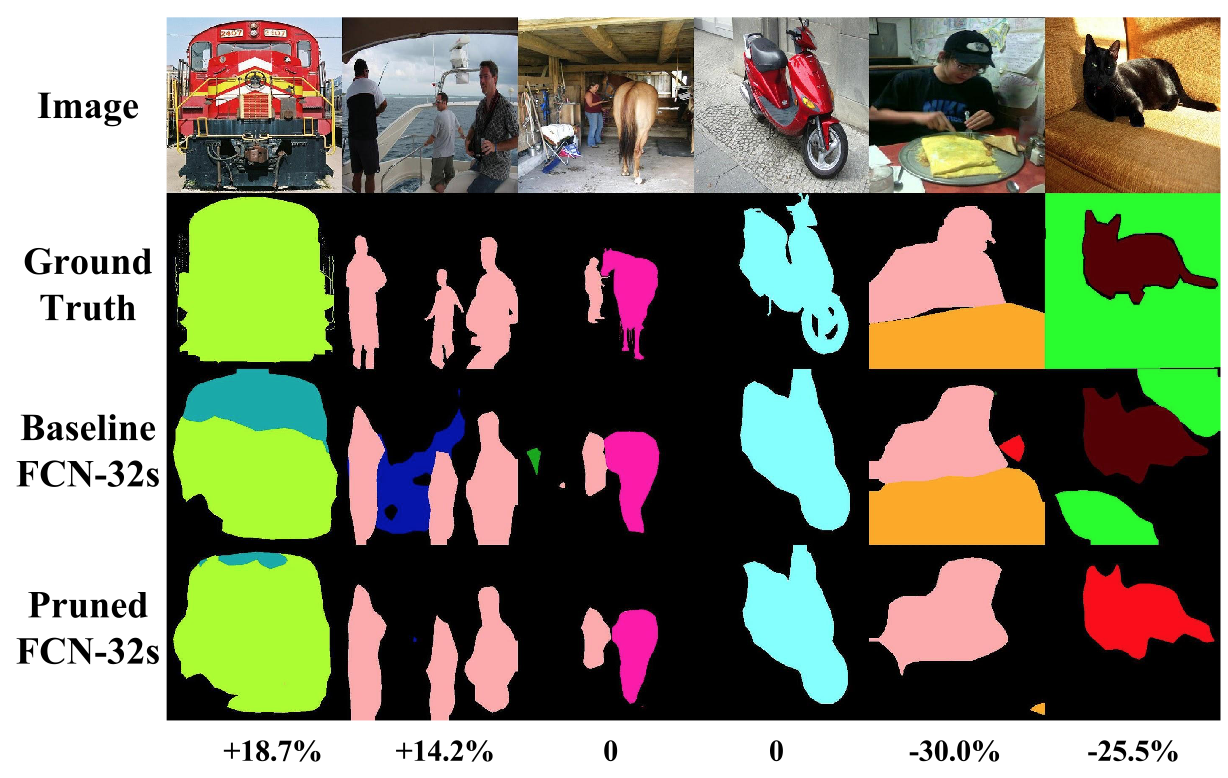}
\end{center}
   \caption{Segmentation visualization of the FCN-32s network on Pascal VOC. The number in each column represents the change of global accuracy. The first two, middle two, and last two columns are samples with global accuracies increased, unchanged, and decreased, respectively.}
\label{fig:change_fcn}
\end{figure*}

We also apply the magnitude based method in \cite{iclr17} to the same baseline using the same pruning ratio. Compared to their method, our algorithm can find a better combination of filters to remove which results in a lower accuracy drop (1.5 vs 3.4). Moreover, some sample segmentation results after pruning are presented and compared with original results in Fig. \ref{fig:change_fcn}.

\subsection{Pruning SegNet on CamVid}
We also experiment with a network with different architecture, SegNet \cite{segnet}, on a different dataset, the CamVid dataset \cite{camvid}. The drop bound is set as $b=2$. Pruning results are reported in Table. \ref{tl:fcn}. Also, the method in \cite{iclr17} is also applied to the same baseline with same pruning ratio for comparison. Our algorithm removes near 56.9\% of parameters in the baseline SegNet and speeds it up by 42.4\% on GPU and 53.0\% on CPU. The global accuracy of the pruned network produced by our method is increased 2.1\% while the network produced by the magnitude-based decreases 3.0\%. This is because our reward function not only guarantees the accuracy not to drop below a specified level but also encourages higher accuracies. The SegNet network is twice larger than its building block, the VGG-16 network. However, CamVid is a small scale dataset. In this case, our pruning algorithm prevents the network from over-fitting which results in a higher global accuracy after pruning.

\begin{figure}[h]
  \centering
  \captionsetup{justification=centering}
  \begin{subfigure}{.5\textwidth}
  \centering
  \includegraphics[width=0.7\linewidth]{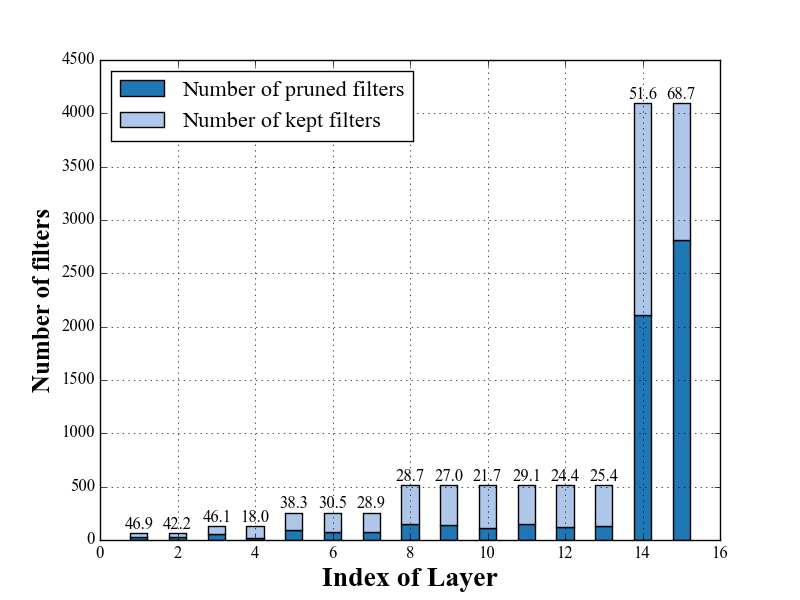}
   \captionof{figure}{Results of pruning the FCN-32s network on the Pascal VOC dataset.}
  \label{fig:ratio_fcn}
\end{subfigure}%
\hfill
\begin{subfigure}{.5\textwidth}
  \centering
  \includegraphics[width=0.7\linewidth]{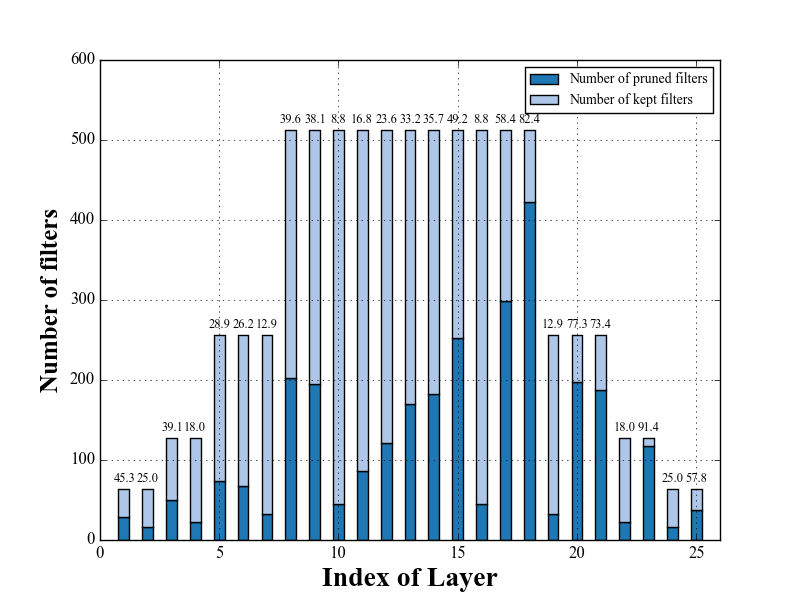}
   \captionof{figure}{Results of pruning the SegNet network on the CamVid dataset.}
  \label{fig:ratio_segnet}
\end{subfigure}%
\caption{ Pruning results of CNNs on segmentation tasks. Numbers on top of the bars are the pruning ratios.}
\label{fig:segmentation}
\end{figure}

\begin{figure*}
  \centering
  \captionsetup{justification=centering}
  
  \includegraphics[width=0.8\linewidth]{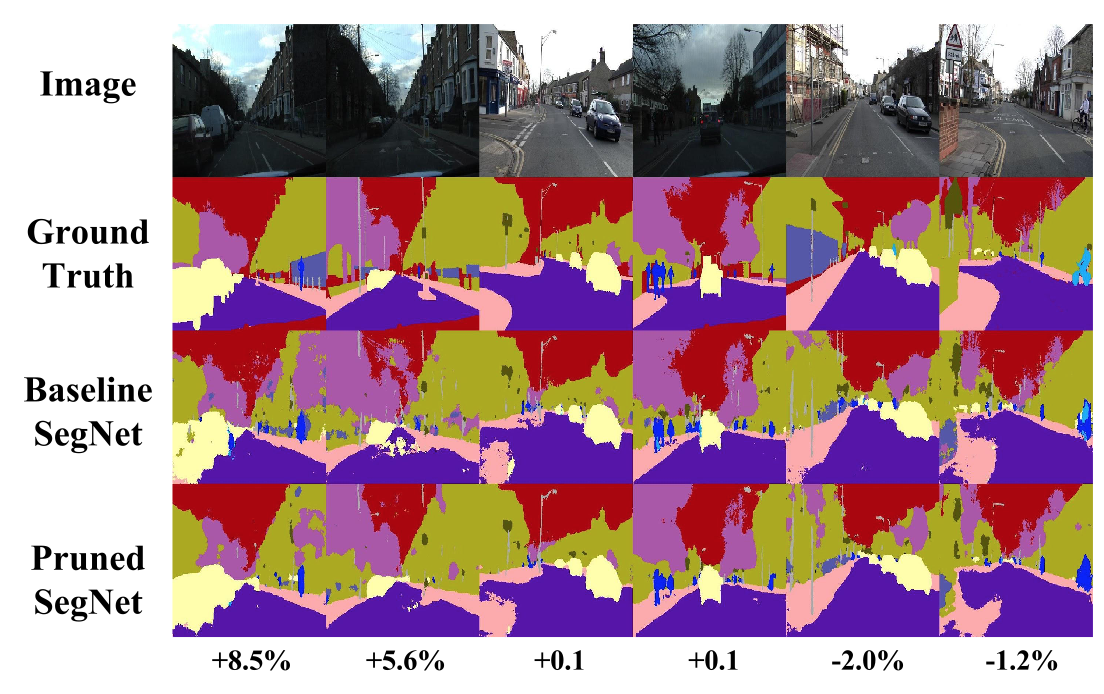}

\caption{Segmentation visualization of the SegNet network on CamVid. The number in each column represents the change of global accuracy. The first two, middle two, and last two columns are samples with global accuracies increased, unchanged, and decreased, respectively.}
\label{fig:change_segnet}
\end{figure*}

Detailed pruning ratios of each layer in the FCN-32s network is presented in Fig. \ref{fig:ratio_segnet}. In the SegNet network, the first half and second half part of the network are symmetrical. However, the pruning results show that the second half contains more unnecessary filters than the first half. Only 26.9\% of the filters are removed in the first half. In contrast, 49.2\% of the filters are removed in the second half. Some segmentation visualizations of the SegNet network on CamVid are presented in Fig. \ref{fig:change_segnet}.

\section{Conclusion}

This paper introduces a ``try-and-learn" learning algorithm for pruning filters in convolutional neural networks. Our work focuses on the following questions: 1). how to prune redundant CNN filters in a data-driven way; 2). how to enable the control of the tradeoff between network performance and its scale in pruning. Our training algorithm is based the policy gradient method. By using a novel reward function, our method aggressively prunes the filters in baseline network while maintaining the performance in a desired level. We benchmark our algorithm on several widely used visual recognition and semantic segmentation CNN networks. Experimental results demonstrate the effectiveness of our algorithm. Potential future directions of our work include 1). extending our method to a more efficient learning algorithm to reduce the training time; 2). formulating the pruning of the entire network as one learning task for higher automation.

{\small
\bibliographystyle{ieee}
\bibliography{egbib}

\begin{thebibliography}{10}\itemsep=-1pt

\bibitem{alvarez2016learning}
J.~M. Alvarez and M.~Salzmann.
\newblock Learning the number of neurons in deep networks.
\newblock In {\em Advances in Neural Information Processing Systems}, pages
  2270--2278, 2016.

\bibitem{segnet}
V.~Badrinarayanan, A.~Kendall, and R.~Cipolla.
\newblock Segnet: A deep convolutional encoder-decoder architecture for image
  segmentation.
\newblock {\em arXiv preprint arXiv:1511.00561}, 2015.

\bibitem{nd}
B.~Baker, O.~Gupta, N.~Naik, and R.~Raskar.
\newblock Designing neural network architectures using reinforcement learning.
\newblock {\em International Conference on Learning Representations}, 2017.

\bibitem{camvid}
G.~J. Brostow, J.~Fauqueur, and R.~Cipolla.
\newblock Semantic object classes in video: A high-definition ground truth
  database.
\newblock {\em Pattern Recognition Letters}, 30(2):88--97, 2009.

\bibitem{ppdl}
M.~Denil, B.~Shakibi, L.~Dinh, N.~de~Freitas, et~al.
\newblock Predicting parameters in deep learning.
\newblock In {\em Advances in Neural Information Processing Systems}, pages
  2148--2156, 2013.

\bibitem{exploit_linear}
E.~L. Denton, W.~Zaremba, J.~Bruna, Y.~LeCun, and R.~Fergus.
\newblock Exploiting linear structure within convolutional networks for
  efficient evaluation.
\newblock In {\em Advances in Neural Information Processing Systems}, pages
  1269--1277, 2014.

\bibitem{lccl}
X.~Dong, J.~Huang, Y.~Yang, and S.~Yan.
\newblock More is less: A more complicated network with less inference
  complexity.
\newblock In {\em The IEEE International Conference on Computer Vision (ICCV)},
  October 2017.

\bibitem{pascal}
M.~Everingham, L.~Van~Gool, C.~Williams, J.~Winn, and A.~Zisserman.
\newblock The pascal visual object classes challenge 2012 (voc2012) results
  (2012).
\newblock In {\em URL http://www. pascal-network.
  org/challenges/VOC/voc2011/workshop/index. html}, 2010.

\bibitem{dns}
Y.~Guo, A.~Yao, and Y.~Chen.
\newblock Dynamic network surgery for efficient dnns.
\newblock In {\em Advances In Neural Information Processing Systems}, pages
  1379--1387, 2016.

\bibitem{eie}
S.~Han, X.~Liu, H.~Mao, J.~Pu, A.~Pedram, M.~A. Horowitz, and W.~J. Dally.
\newblock Eie: efficient inference engine on compressed deep neural network.
\newblock In {\em Proceedings of the 43rd International Symposium on Computer
  Architecture}, pages 243--254. IEEE Press, 2016.

\bibitem{han_deep}
S.~Han, H.~Mao, and W.~J. Dally.
\newblock Deep compression: Compressing deep neural networks with pruning,
  trained quantization and huffman coding.
\newblock {\em International Conference on Learning Representations}, 2016.

\bibitem{han_both}
S.~Han, J.~Pool, J.~Tran, and W.~Dally.
\newblock Learning both weights and connections for efficient neural network.
\newblock In {\em Advances in Neural Information Processing Systems}, pages
  1135--1143, 2015.

\bibitem{second}
B.~Hassibi, D.~G. Stork, et~al.
\newblock Second order derivatives for network pruning: Optimal brain surgeon.
\newblock {\em Advances in neural information processing systems}, pages
  164--164, 1993.

\bibitem{resnet}
K.~He, X.~Zhang, S.~Ren, and J.~Sun.
\newblock Deep residual learning for image recognition.
\newblock In {\em Proceedings of the IEEE Conference on Computer Vision and
  Pattern Recognition}, pages 770--778, 2016.

\bibitem{yihuihe}
Y.~He, X.~Zhang, and J.~Sun.
\newblock Channel pruning for accelerating very deep neural networks.
\newblock In {\em The IEEE International Conference on Computer Vision (ICCV)},
  Oct 2017.

\bibitem{bnn}
I.~Hubara, M.~Courbariaux, D.~Soudry, R.~El-Yaniv, and Y.~Bengio.
\newblock Binarized neural networks.
\newblock In D.~D. Lee, M.~Sugiyama, U.~V. Luxburg, I.~Guyon, and R.~Garnett,
  editors, {\em Advances in Neural Information Processing Systems 29}, pages
  4107--4115. Curran Associates, Inc., 2016.

\bibitem{lr_expansion}
M.~Jaderberg, A.~Vedaldi, and A.~Zisserman.
\newblock Speeding up convolutional neural networks with low rank expansions.
\newblock {\em CoRR}, abs/1405.3866, 2014.

\bibitem{caffe}
Y.~Jia, E.~Shelhamer, J.~Donahue, S.~Karayev, J.~Long, R.~Girshick,
  S.~Guadarrama, and T.~Darrell.
\newblock Caffe: Convolutional architecture for fast feature embedding.
\newblock {\em arXiv preprint arXiv:1408.5093}, 2014.

\bibitem{adam}
D.~Kingma and J.~Ba.
\newblock Adam: A method for stochastic optimization.
\newblock {\em International Conference on Learning Representations}, 2014.

\bibitem{cifar}
A.~Krizhevsky and G.~Hinton.
\newblock Learning multiple layers of features from tiny images.
\newblock 2009.

\bibitem{alexnet}
A.~Krizhevsky, I.~Sutskever, and G.~E. Hinton.
\newblock Imagenet classification with deep convolutional neural networks.
\newblock In {\em Advances in neural information processing systems}, pages
  1097--1105, 2012.

\bibitem{optimal}
Y.~LeCun, J.~S. Denker, and S.~A. Solla.
\newblock Optimal brain damage.
\newblock In {\em Advances in neural information processing systems}, pages
  598--605, 1990.

\bibitem{iclr17}
H.~Li, A.~Kadav, I.~Durdanovic, H.~Samet, and H.~P. Graf.
\newblock Pruning filters for efficient convnets.
\newblock {\em International Conference on Learning Representations}, 2017.

\bibitem{fcn}
J.~Long, E.~Shelhamer, and T.~Darrell.
\newblock Fully convolutional networks for semantic segmentation.
\newblock In {\em Proceedings of the IEEE Conference on Computer Vision and
  Pattern Recognition}, pages 3431--3440, 2015.

\bibitem{xnor}
M.~Rastegari, V.~Ordonez, J.~Redmon, and A.~Farhadi.
\newblock Xnor-net: Imagenet classification using binary convolutional neural
  networks.
\newblock In {\em European Conference on Computer Vision}, pages 525--542.
  Springer, 2016.

\bibitem{Q2}
D.~Silver, A.~Huang, C.~J. Maddison, A.~Guez, L.~Sifre, G.~Van Den~Driessche,
  J.~Schrittwieser, I.~Antonoglou, V.~Panneershelvam, M.~Lanctot, et~al.
\newblock Mastering the game of go with deep neural networks and tree search.
\newblock {\em Nature}, 529(7587):484--489, 2016.

\bibitem{vgg}
K.~Simonyan and A.~Zisserman.
\newblock Very deep convolutional networks for large-scale image recognition.
\newblock {\em arXiv preprint arXiv:1409.1556}, 2014.

\bibitem{BMVC2015_31}
S.~Srinivas and R.~V. Babu.
\newblock Data-free parameter pruning for deep neural networks.
\newblock {\em arXiv preprint arXiv:1507.06149}, 2015.

\bibitem{pg}
R.~S. Sutton, D.~A. McAllester, S.~P. Singh, Y.~Mansour, et~al.
\newblock Policy gradient methods for reinforcement learning with function
  approximation.
\newblock In {\em NIPS}, volume~99, pages 1057--1063, 1999.

\bibitem{googlenet}
C.~Szegedy, W.~Liu, Y.~Jia, P.~Sermanet, S.~Reed, D.~Anguelov, D.~Erhan,
  V.~Vanhoucke, and A.~Rabinovich.
\newblock Going deeper with convolutions.
\newblock In {\em Proceedings of the IEEE Conference on Computer Vision and
  Pattern Recognition}, pages 1--9, 2015.

\bibitem{Q1}
C.~J. C.~H. Watkins.
\newblock {\em Learning from delayed rewards}.
\newblock PhD thesis, University of Cambridge England, 1989.

\bibitem{Wen_2017_ICCV}
W.~Wen, C.~Xu, C.~Wu, Y.~Wang, Y.~Chen, and H.~Li.
\newblock Coordinating filters for faster deep neural networks.
\newblock In {\em The IEEE International Conference on Computer Vision (ICCV)},
  October 2017.

\bibitem{reinforce}
R.~J. Williams.
\newblock Simple statistical gradient-following algorithms for connectionist
  reinforcement learning.
\newblock {\em Machine learning}, 8(3-4):229--256, 1992.

\bibitem{zhang_accelerating}
X.~Zhang, J.~Zou, K.~He, and J.~Sun.
\newblock Accelerating very deep convolutional networks for classification and
  detection.
\newblock {\em IEEE transactions on pattern analysis and machine intelligence},
  38(10):1943--1955, 2016.

\bibitem{zhou2016less}
H.~Zhou, J.~M. Alvarez, and F.~Porikli.
\newblock Less is more: Towards compact cnns.
\newblock In {\em European Conference on Computer Vision}, pages 662--677.
  Springer, 2016.

\bibitem{ns}
B.~Zoph and Q.~V. Le.
\newblock Neural architecture search with reinforcement learning.
\newblock {\em arXiv preprint arXiv:1611.01578}, 2016.

\end{thebibliography}
}

\end{document}